%% file: root.tex
\documentclass[letterpaper, 10 pt, conference]{ieeeconf}  % Comment this line out if you need a4paper

\IEEEoverridecommandlockouts                              % This command is only needed if 
\usepackage{graphicx} % for pdf, bitmapped graphics files
\usepackage{amsmath} % assumes amsmath package installed
\usepackage{amssymb}  % assumes amsmath package installed
\usepackage{xcolor}
\usepackage{cite}
\usepackage{booktabs}
\usepackage{dblfloatfix}

\usepackage{etoolbox}
\makeatletter
\patchcmd{\@makecaption}
  {\scshape}
  {}
  {}
  {}
\makeatother

\title{\LARGE \bf
Predicting Dynamic Map States from Limited Field-of-View Sensor Data
}
\author{Knut Peterson$^{1}$ and David Han$^{1}$% <-this % stops a space
\thanks{$^{1}$iMaPLe Research Lab, Drexel University, Philadelphia, PA}
\thanks{\tt \{kp3275, dkh42\}@drexel.edu}
}

\begin{document}

\maketitle
\thispagestyle{empty}
\pagestyle{empty}

%%%%%%%%%%%%%%%%%%%%%%%%%%%%%%%%%%%%%%%%%%%%%%%%%%%%%%%%%%%%%%%%%%%%%%%%%%%%%%%%
\begin{abstract}
When autonomous systems are deployed in real-world scenarios, sensors are often subject to limited field-of-view (FOV) constraints, either naturally through system design, or through unexpected occlusions or sensor failures. In conditions where a large FOV is unavailable, it is important to be able to infer information about the environment and predict the state of nearby surroundings based on available data to maintain safe and accurate operation. In this work, we explore the effectiveness of deep learning for dynamic map state prediction based on limited FOV time series data. We show that by representing dynamic sensor data in a simple single-image format that captures both spatial and temporal information, we can effectively use a wide variety of existing image-to-image learning models to predict map states with high accuracy in a diverse set of sensing scenarios.

\end{abstract}

%%%%%%%%%%%%%%%%%%%%%%%%%%%%%%%%%%%%%%%%%%%%%%%%%%%%%%%%%%%%%%%%%%%%%%%%%%%%%%%%
\input{1.Introduction}

\input{2.Related}

\input{3.Proposed}

\input{4.Experiment}

\input{5.Conclusion}


%%%%%%%%%%%%%%%%%%%%%%%%%%%%%%%%%%%%%%%%%%%%%%%%%%%%%%%%%%%%%%%%%%%%%%%%%%%%%%%%

%%%%%%%%%%%%%%%%%%%%%%%%%%%%%%%%%%%%%%%%%%%%%%%%%%%%%%%%%%%%%%%%%%%%%%%%%%%%%%%%



% \section*{ACKNOWLEDGMENT}
% The funding for this work was provided by \textcolor{red}{TODO}

%%%%%%%%%%%%%%%%%%%%%%%%%%%%%%%%%%%%%%%%%%%%%%%%%%%%%%%%%%%%%%%%%%%%%%%%%%%%%%%%

\bibliographystyle{IEEEtran}
\bibliography{ref.bib}
\end{document}

%% file: 1.Introduction.tex
\section{Introduction}

Humans eyesight is restricted to a relatively narrow field of view (FOV), but even with a limited FOV we manage to maintain a surprisingly accurate mental model of our surroundings. Even more impressively, we can track and predict the movements of objects, people, or vehicles that pass through our field of view for only a brief time. As automated systems continue to increase in popularity and usefulness through the widespread adoption of self-driving cars and mobile robots, it is important to consider how automated systems can achieve the same level of prediction and tracking for objects outside their current FOV. In some cases, sensor FOV is limited by platform constraints, and even in cases where large FOV sensing is feasible, external factors can still cause occlusions and impact available sensor information. By making use of limited information from sensors to predict the states of objects that have left the sensor FOV or are temporarily blocked from view, we can improve the safety and performance of autonomous systems.

Although tracking and prediction of moving obstacle states using limited FOV sensors has previously been explored, it has been mainly analyzed through the creation of statistical models \cite{doherty2017, sung2017searching, catellani2023probabilistic} or circumvented through distributed methods to achieve a large effective FOV by combining the sensing capabilities of multiple agents \cite{mukherjee2022distributed, nguyen2021distributed}. Recently, however, more work has been done to approach the problem of state prediction, especially as the rise of deep learning has shown how data-driven approaches can be very effective for many kinds of predictive tasks. This is mainly because they reduce the need for complex handcrafted models and instead focus on the collection or creation of large amounts of training data and letting a deep learning model learn trends directly from the data. Deep learning models have already been used for map prediction and navigation with effective results \cite{katyal2019uncertainty, katyal2018occupancy, katyal2021navigation, xie2023sogmp, xia2024, sharma2023, hoermann2018dynamic, guizilini2019hilbert}, but little work has been done to explore this approach to limited FOV problems.
By exploring methods for applying deep-learning approaches to map state prediction from limited FOV sensor data, we hope to increase the effectiveness of agents in real world scenarios where perfect environmental sensing is not possible. In this work, we explore an approach to representing dynamic limited FOV sensor data in a simple single-image format that captures both spatial and temporal information by representing LIDAR scan points as time-decaying gray-scale pixel intensities in a map image. This allows sequential sensor information to be captured in a single image, and allows it to be easily processed using existing image-to-image prediction models.

We analyze the effectiveness of a variety of models through the generation of datasets covering four different scenarios of obstacle and robot movement patterns, capturing both limited FOV sensor data and ground truth map states to train and evaluate predictions. Through experiments utilizing these datasets, we show that existing image-to-image learning methods can be effectively applied to limited FOV map prediction, and explore what factors make this approach successful. Specifically, our contributions are as follows.
\begin{itemize}
    \item A unique method of cumulative dynamic sensor projection to represent dynamic limited FOV time series data to enable map prediction using well-established models.
    \item Experimental results using a range of current segmentation models showing the generalization ability of this method for limited-FOV map state prediction.
\end{itemize}

%% file: 2.Related.tex
\section{Related Work}

\begin{figure*}[t!]
    \centering
    \centerline{\includegraphics[width=1\textwidth]{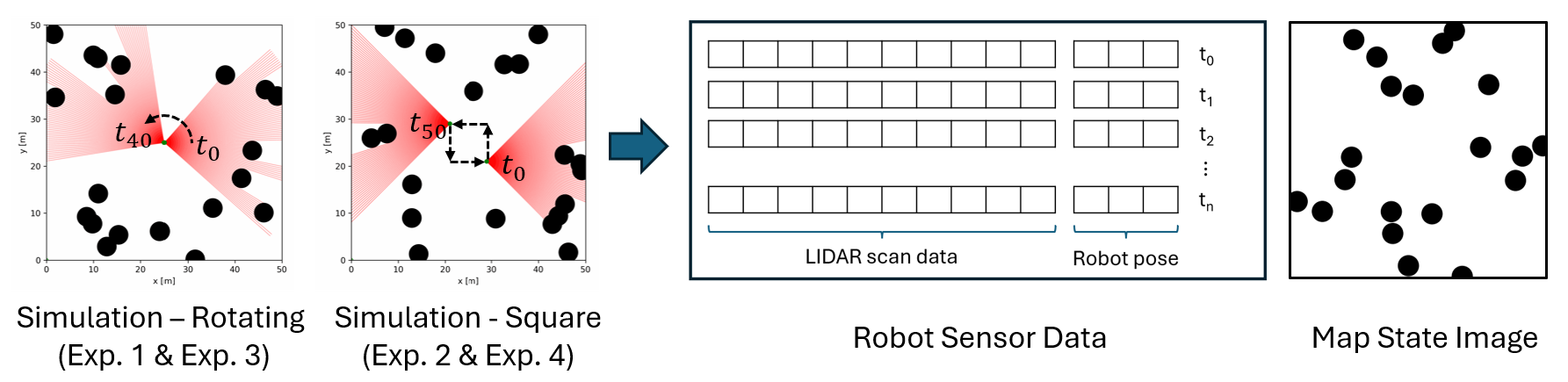}}
    \caption{During data collection, the robot operates in a 2D world of static (exp1 and exp2) or dynamic (exp3 and exp4) obstacles, and gathers data with a LIDAR sensor by rotating (exp1 and exp3) or moving around the world in a square shape (exp2 and exp4). Each run consists of 100 time steps. For each time step the LIDAR scan data and robot pose are recorded, and the map state image is saved at the end of each run.}
    \label{fig:data_collection}
\end{figure*}

\subsection{Limited FOV Sensors}

The problem of effectively utilizing information from limited FOV sensors for predicting or tracking object states has been explored in a variety of contexts, each of which comes with a separate array of challenges.

Sung et. al \cite{sung2017searching} investigated the problem of limited FOV object tracking through a framework based on Bayesian Random Finite Sets, and created a generalized version of the Gaussian Mixture Probability Hypothesis Density filter to be used for both search and tracking. This approach proved effective for tracking variable numbers of targets, in addition to predicting nonlinear target trajectories. Nguyen et. al. \cite{nguyen2021distributed} approached limited FOV tracking from a different angle, where the tracking problem is distributed between multiple nodes, each containing a limited FOV sensor. Other approaches, such as Mukherjee et. al. \cite{mukherjee2022distributed}, focused less on the problem of tracking objects that move out of sensor FOVs, but instead on the control problem of keeping targets within view. In \cite{catellani2023probabilistic}, Catellani et. al. combined several of the previous problems by developing an approach for multi-robot control in communication-denied scenarios, requiring each robot to track nearby agents and accomplish an area mapping task. This was accomplished through a particle filter method for approximating robot states that were outside of the current robot view, which were then shaped by new sensor inputs.

% Other applications of limited FOV sensing include problems of path planning and mapping, especially using UAVs, such as Eaton et. al. \cite{eaton2017uavplanning} who explored the use of Partially Observable Markov Decision Processes (POMDP) for planning paths using limited FOV sensors, Zhao et. al. \cite{zhao2024uavmapping} who explored adaptive planning strategies for mapping, and Lu et. al. \cite{lu2022} who explored the flight control problem using limited FOV sensor information.

\subsection{Occupancy Map Prediction}

Map prediction methods have also been widely explored, with methods ranging from statistical inference to deep learning prediction. Doherty et. al. \cite{doherty2017} explored using Bayesian generalized kernel inference for predicting occupancy maps with effective results. In \cite{katyal2019uncertainty}, Katyal et. al. explored the use of Generative Adversarial Networks (GANs) for predicting occupancy map states based on limited range data. By using GAN-generated expanded map predictions, they were able to apply an information-theoretic exploration strategy based on the variances in generated results. In a follow on paper \cite{katyal2021navigation}, they also explored the potential uses of map prediction methods for enabling high-speed navigation and collision-free planning. In another approach to map prediction, Xie et. al \cite{xie2023sogmp} utilized variational auto-encoders to generate ranges of possible future environment states based on existing knowledge of robot motion, dynamic obstacle states, and static object geometry. Shrestha et. al. \cite{shrestha2019} also used variational auto-encoders for map prediction, however they focus on map in-painting for prediction, and apply their results to improving exploration strategies through region exploration cost prediction.
Other deep learning approaches include Hoermann et. al. \cite{hoermann2018dynamic} who created a CNN based model for autonomous driving occupancy grid prediction, Xia et. al. \cite{xia2024} who explored using U-Nets for predicting occupancy maps for dynamic motion planning, and Sharma et. al. \cite{sharma2023} who developed Proxmap as an alternative model that outperformed U-Net and GAN baselines.
% Other works, such as Vacek et. al. \cite{vacek2022intensities} and Yang et. al. \cite{yang2023vidar} explore the potential of predicting sensor values directly from previous observations, mainly through the prediction of future LIDAR reading values instead of predicting full map states.

While a variety of methods have been explored for limited FOV state prediction, many methods are based around statistical modeling approaches that require complex calculations or large numbers of particles for tracking, which do not scale well to many objects. Additionally, for deep-learning methods, much work has been done in predicting static maps from limited data, and for predicting dynamic map states based on full environmental understanding.  However, little work has been done for predicting dynamic map states using limited FOV data. In this work we hope to show that deep-learning methods can be effectively used for this task.

%% file: 3.Proposed.tex
\section{Proposed Method}

\subsection{Method Overview}

\begin{figure*}[t!]
    \centering
    \centerline{\includegraphics[width=1\textwidth]{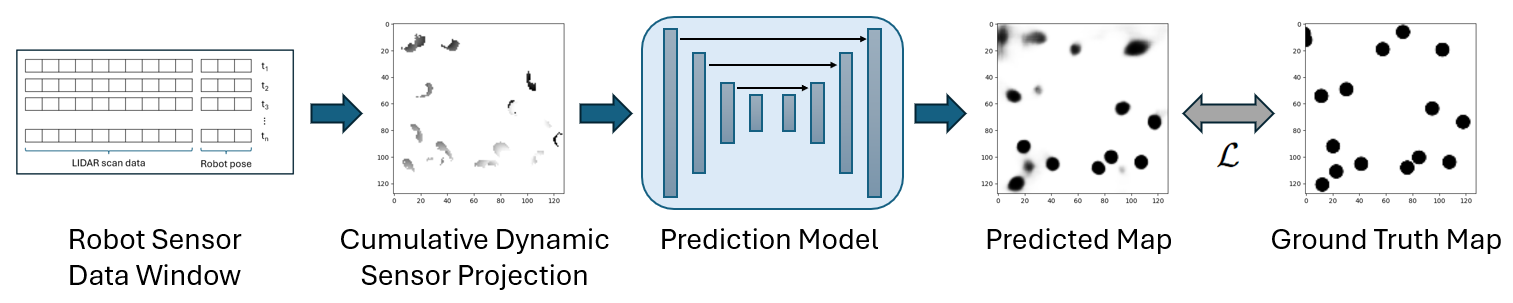}}
    \caption{An overview of our prediction method. A time window of collected LIDAR sensor data and robot pose information is first transformed into a single image using our proposed method of cumulative dynamic sensor projection. That image is then used as the input to an image-to-image prediction model which predicts a final map state based on the sensor input, and the predicted map is compared to the ground truth map state for training and evaluation.}
    \label{fig:architecture}
\end{figure*}

Figure \ref{fig:architecture} shows an overview of our proposed method. First, a time window of collected LIDAR sensor data and robot pose information is transformed into a single image using our proposed cumulative dynamic sensor projection method. That image is then used as the input to an image-to-image prediction model which predicts a final map state based on the sensor input, and the predicted map is compared to the ground truth map state for training and evaluation.

\subsection{Simulation Approach}

To generate data, we created a 2D simulation environment using the IR-SIM Python simulator \cite{irsim} with randomly placed obstacles and a robot equipped with a limited FOV LIDAR sensor. Through four different experimental setups, we changed the dynamics of both the obstacles and robot in order to explore the impact of different levels of sensor information on the the quality of predicted map states.

Figure \ref{fig:data_collection} shows an overview of the data collection method. For each run, the simulation was run for a total of 100 time steps, with the robot sensor data and pose being saved at every time step, and the ground truth map state image at the final time step was saved at the end of the run. The simulation was then repeated with the robot executing the rotation or movement path in the opposite direction. For the simulation, the robot LIDAR sensor was set to a 90 degree FOV, a ray count of 100, and a maximum range of 30m. The map was set to a size of 50x50m.

For experiment 1 and experiment 2, the obstacles were placed randomly in the environment and remained static throughout each run. For experiment 3 and experiment 4, the obstacles were again randomly initialized but moved in straight lines at constant speeds, changing directions randomly at specified intervals throughout the run, or whenever they entered the center zone occupied by the robot to prevent collisions that would fully occlude the LIDAR sensor.

For experiment 1 and experiment 3, the robot did not move around the map but simply rotated in place, completing one full rotation in the 100 time step prediction cycle. For experiment 2 and experiment 4, the robot moved in a square shape in the center of the map. For each of the four experiments, 10k pairs of LIDAR data and ground truth map images were generated as a training set, an additional 128 pairs as a validation set, and 500 pairs as a test set.

\subsection{Cumulative Dynamic Sensor Projection}

Learning from raw sensor data is difficult, and existing models are not suited to learning abstract spatial relationships with limited information. Therefore, in order to convert the data into a more useful format, we pre-process the scan points into a global coordinate frame using the robot pose, LIDAR range, and angle values. Each LIDAR scan produces a vector of distance values, as illustrated in Figure \ref{fig:architecture}. While these vectors can be directly fed into a network to form a 2D obstacle map, it would be more effective to convert them into a more spatially meaningful representation. To achieve this, we project the obstacle boundary points from the LIDAR point clouds onto a 2D map using the robot pose. % These projections are accumulated progressively as additional LIDAR inputs are converted into the 2D map, enhancing its accuracy over time.

\begin{figure}[t]
    \centering
    \centerline{\includegraphics[width=0.92\linewidth]{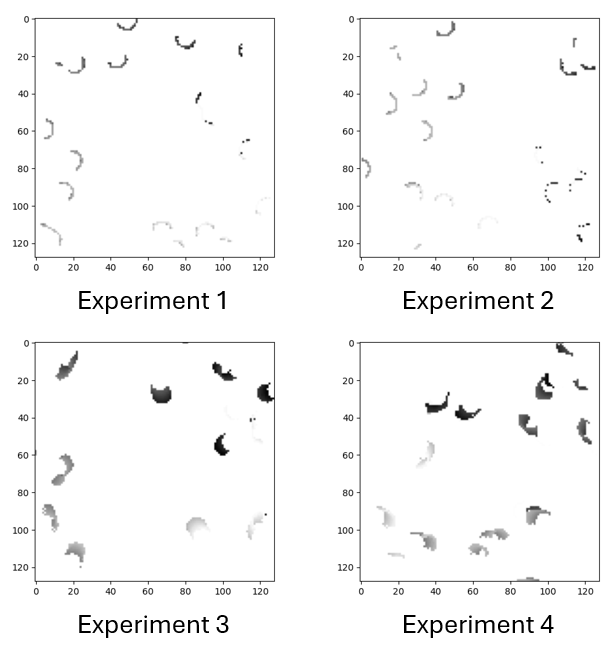}}
    \caption{By representing limited FOV LIDAR scan data as time-decaying gray scale pixel intensities, we capture both spatial and temporal information. In experiments with static obstacles (exp1 and exp2) we see detailed obstacle outlines, with decaying pixel intensities based on when they were last sensed. In experiments with dynamic obstacles (exp3 and exp4) we see object motion recorded through pixel intensity gradients.}
    \label{fig:data_representation}
\end{figure}

Additionally, we need to differentiate LIDAR scans from across a window of time steps. One method of doing this is to use a series of map state images as a model input, allowing it to learn obstacle movement from changes between images. However, this requires storing many images for training, and using multiple images or channels adds additional model complexity and size. Instead, we propose a more elegant solution of capturing the cumulative sensor readings from a window of time steps into a single image, and differentiating them by time-decaying gray-scale intensity. This results in a single image representing a window of sensor values, with darker values representing recent sensor readings, and with lighter values representing older sensor readings, effectively capturing obstacle dynamics with gray-scale gradients. Figure \ref{fig:data_representation} shows example processed sensor images for each of the four experiments. 

From the figure, we can notice a few characteristics that emerge as a result of this method of data representation. First, the experiments with static obstacles have relatively well-defined edges where obstacles are sensed, and there is minimal pixel intensity decay visible as older sensor values are overridden by newer ones. However, in the experiments with moving obstacles, we can easily see how obstacle locations change over time based on the increasing pixel intensities across time steps, forming natural gradients with object motion. We can also note that some areas of the image only have very low pixel intensities as they have not been scanned recently, while other areas are much darker, corresponding to more recent sensor readings.

\begin{table*}[!tb]
% \footnotesize
\centering
\caption{Prediction metrics across all four experiments for all models. The results for most metrics are high across experiments and models, however there is a noticeable drop particularly in SSIM and Specificity between Experiment 2 and Experiment 3 as obstacles change from stationary to dynamic.}
\label{table:results}
\begin{tabular}{@{}lllllllllllll@{}}
\toprule
\multicolumn{1}{c|}{\textbf{STATIC}} &
  \multicolumn{6}{c|}{\textbf{Experiment 1 - Rotation}} &
  \multicolumn{6}{c}{\textbf{Experiment 2 - Square}} \\ \midrule
\multicolumn{1}{l|}{\textbf{Model}} &
  \textbf{SSIM} &
  \textbf{Acc.} &
  \textbf{Dice} &
  \textbf{Prec.} &
  \textbf{Recall} &
  \multicolumn{1}{l|}{\textbf{Spec.}} &
  \textbf{SSIM} &
  \textbf{Acc.} &
  \textbf{Dice} &
  \textbf{Prec.} &
  \textbf{Recall} &
  \textbf{Spec.} \\ \midrule
\multicolumn{1}{l|}{U-Net} &
  0.7415 &  \textbf{0.9770} &  \textbf{0.9873} &  0.9829 &  0.9917 &
  \multicolumn{1}{l|}{0.8443} &
  0.7358 &  \textbf{0.9858} &  \textbf{0.9921} &  \textbf{0.9901} &  \textbf{0.9942} &  \textbf{0.9103} \\ \midrule
\multicolumn{1}{l|}{FPN} &  \textbf{0.7350} &  0.9746 &  0.9860 &  0.9812 &  0.9908 &
  \multicolumn{1}{l|}{0.8282} &
  \textbf{0.7327} &  0.9830 &  0.9906 &  0.9893 &  0.9919 &  0.9032 \\ \midrule
\multicolumn{1}{l|}{UPerNet} &
  0.7367 &  0.9749 &  0.9861 &  \textbf{0.9832} &  0.9890 &
  \multicolumn{1}{l|}{\textbf{0.8475}} &
  0.7340 &  0.9837 &  0.9909 &  0.9897 &  0.9922 &  0.9073 \\ \midrule
\multicolumn{1}{l|}{Segformer} &
  0.7425 &  0.9755 &  0.9865 &  0.9800 &  \textbf{0.9930} &
  \multicolumn{1}{l|}{0.8168} &
  0.7364 &  0.9827 &  0.9904 &  0.9876 &  0.9932 &  0.8880 \\ \midrule
\multicolumn{1}{l|}{\textbf{Average:}} &
  \textbf{0.7389} &
  \textbf{0.9755} &
  \textbf{0.9865} &
  \textbf{0.9818} &
  \textbf{0.9911} &
  \multicolumn{1}{l|}{\textbf{0.8342}} &
  \textbf{0.7347} &
  \textbf{0.9838} &
  \textbf{0.9910} &
  \textbf{0.9892} &
  \textbf{0.9929} &
  \textbf{0.9022} \\ \midrule
 &   &   &   &   &   &   &   &   &   &   &   &
   \\ \midrule
\multicolumn{1}{c|}{\textbf{DYNAMIC}} &
  \multicolumn{6}{c|}{\textbf{Experiment 3 - Rotation}} &
  \multicolumn{6}{c}{\textbf{Experiment 4 - Square}} \\ \midrule
\multicolumn{1}{l|}{\textbf{Model}} &
  \textbf{SSIM} &
  \textbf{Acc.} &
  \textbf{Dice} &
  \textbf{Prec.} &
  \textbf{Recall} &
  \multicolumn{1}{l|}{\textbf{Spec.}} &
  \textbf{SSIM} &
  \textbf{Acc.} &
  \textbf{Dice} &
  \textbf{Prec.} &
  \textbf{Recall} &
  \textbf{Spec.} \\ \midrule
\multicolumn{1}{l|}{U-Net} &
  0.6633 &  0.9481 &  0.9719 &  \textbf{0.9638} &  0.9802 &
  \multicolumn{1}{l|}{\textbf{0.5950}} &
  0.6609 &  0.9470 &  0.9713 &  0.9622 &  0.9806 &  0.5830 \\ \midrule
\multicolumn{1}{l|}{FPN} &
  0.6633 &  0.9497 &  0.9729 &  0.9615 &  0.9846 &
  \multicolumn{1}{l|}{0.5666} &
  0.6620 &  \textbf{0.9477} &  \textbf{0.9718} &  0.9605 &  \textbf{0.9833} &  0.5626 \\ \midrule
\multicolumn{1}{l|}{UPerNet} &
  \textbf{0.6774} &  \textbf{0.9497} &  \textbf{0.9731} &  0.9563 &  \textbf{0.9905} &
  \multicolumn{1}{l|}{0.5019} &
  \textbf{0.6531} &  0.9474 &  0.9715 &  \textbf{0.9631} &  0.9800 &  \textbf{0.5939} \\ \midrule
\multicolumn{1}{l|}{Segformer} &  0.6693 &  0.9474 &  0.9717 &  0.9585 &  0.9854 &
  \multicolumn{1}{l|}{0.5305} &
  0.6583 &  0.9447 &  0.9702 &  0.9588 &  0.9818 &  0.5434 \\ \midrule
\multicolumn{1}{l|}{\textbf{Average:}} &
  \textbf{0.6683} &
  \textbf{0.9487} &
  \textbf{0.9724} &
  \textbf{0.9600} &
  \textbf{0.9852} &
  \multicolumn{1}{l|}{\textbf{0.5485}} &
  \textbf{0.6586} &
  \textbf{0.9467} &
  \textbf{0.9712} &
  \textbf{0.9612} &
  \textbf{0.9814} &
  \textbf{0.5707} \\ \bottomrule
\end{tabular}
\end{table*}

\subsection{Prediction Models}

Because we represent the dynamic sensor data with a single image, we can easily apply existing model structures directly to predict map states. We evaluate our data representation approach using the U-Net \cite{unet}, FPN \cite{fpn}, UPerNet \cite{upernet}, and Segformer \cite{xie2021segformer} models, all with the mit\_b0 \cite{xie2021segformer} backbone. We use the model implementations from Segmentation Models \cite{segmodels}, and evaluate with metrics from Piqa \cite{piqa} and Segmetrics \cite{segmetrics}.

For all four experiments we train the model using the Adam optimizer, a learning rate of 1e-4, and early-stopping with a threshold of 0.001 and patience of 5 epochs. For the training loss function we use a combination of Binary Cross-Entropy and Dice loss. Binary Cross-Entropy loss is often used in prediction and segmentation tasks because it is a stable measure of differences between predicted and ground truth binary labels. In the case of map prediction, it ensures that the final results maintain close pixel-wise accuracy to the ground truth.

\begin{equation}
\mathcal{L}_{\text{BCE}} = - \frac{1}{N} \sum_{i=1}^{N} \left[ g_i \log(p_i) + (1 - g_i) \log(1 - p_i) \right]
\end{equation}

Dice Loss is a loss function based on the Dice Coefficient, which measures the overlap between predicted and ground truth regions. This makes it more effective for tasks such as image segmentation or binary map prediction, where there are imbalances in class distributions. It also encourages global alignment of shapes over pixel-wise accuracy, making it particularly useful in map prediction.

\begin{equation}
\mathcal{L}_{\text{Dice}} = 1 - \frac{2 \sum_{i=1}^{N} p_i g_i}{\sum_{i=1}^{N} p_i^2 + \sum_{i=1}^{N} g_i^2 + \epsilon}
\end{equation}

For our final loss function we use the simple combination of both BCE loss and Dice loss to take advantage of both of their strengths in predicting final map states.

\begin{equation}
\mathcal{L} =  \mathcal{L}_{\text{BCE}} + \mathcal{L}_{\text{Dice}}
\end{equation}

%% file: 4.Experiment.tex
\section{Experimental Results}

\subsection{Quantitative Results}

Table \ref{table:results} shows the prediction metrics across the four experimental setups. For the two experiments with static obstacles, we see that experiment 2 had slightly better results than experiment 1. This reflects how with static obstacles, the rotational movement in experiment 1 captured less overall map information than the square movement in experiment 2, as more movement increased sensing coverage to reduce occlusions and improved overall map prediction accuracy.

For the two experiments with dynamic obstacles, we see that experiment 3 had slightly higher metrics than experiment 4. Between the two experiments, the regions located behind the robot as it completed the edges of the square had a larger delay between sensing and prediction than for the case where the robot rotated in place. Because the obstacles were moving, this increased delay from sensing to prediction reduced the accuracy of predictions, showing that with dynamic obstacles the recency of data is more important than long term map coverage.

\begin{figure*}[tbp]
    \centering
    \centerline{\includegraphics[width=0.95\textwidth]{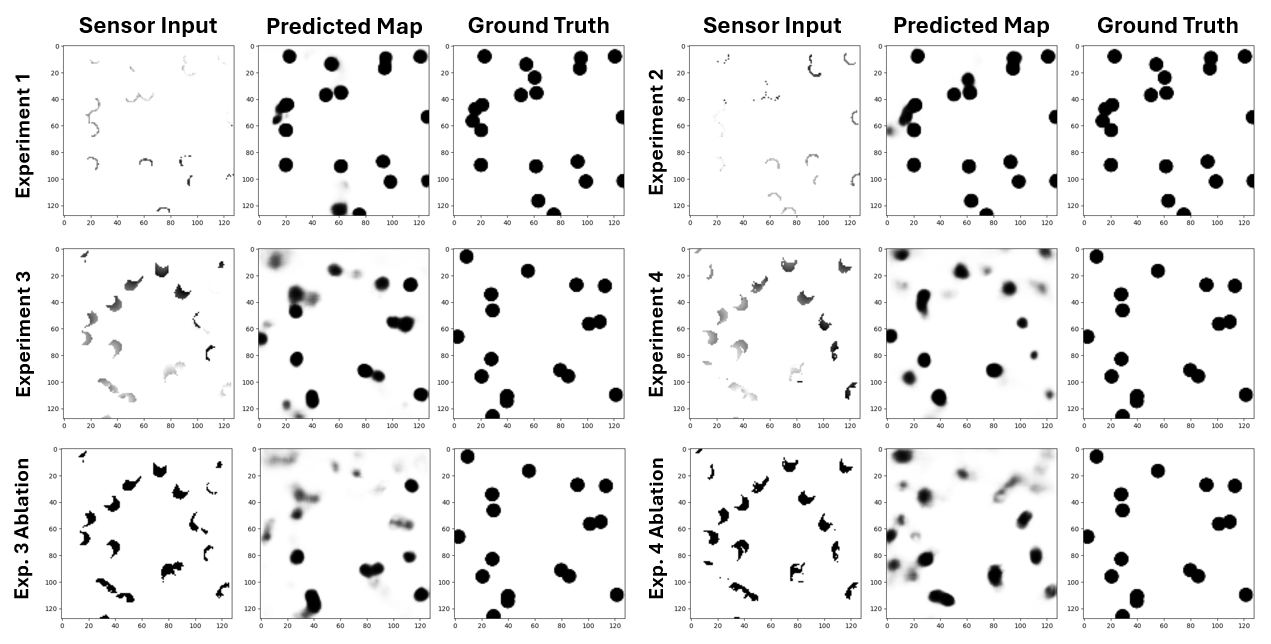}}
    \caption{Example results from the U-Net model for all four experimental setups. Model predictions for experiments with static obstacles (exp. 1 and exp. 2) are quite precise, but miss a few predictions due to occlusions. Results for experiments with dynamic obstacles (exp. 3 and exp. 4) are less precise, with obstacle locations becoming blurry and probabilistic as sensor values become older and less reliable, or as obstacle movement produces additional occlusions. For the ablation results, we see that removing the gray-scale time decay of sensor information negatively impacts the predicted map states. }
    \label{fig:results}
\end{figure*}

Between the first two experiments with static obstacles and the last two with dynamic obstacles, we see a slight drop in all metrics, reflecting the increased difficulty of tracking moving objects. Most notably, SSIM and Specificity dropped substantially compared to the other metrics. Because SSIM measures the preservation of fine-grained structural details, we can infer that the introduction of moving obstacles caused predictions to lose detail in favor of overall accuracy. Similarly, a drop in Specificity reflects a tendency to over-predict positives, likely resulting from the predicted object boundaries becoming stretched to accommodate the uncertainty in their movement. This can be seen clearly in the shift in predictions shown in the Qualitative Results section.

\subsection{Qualitative Results}

Figure \ref{fig:results} shows a set of sensor inputs, predicted map states, and ground truth map states for U-Net model (the best performing model by a small margin) for all four experimental setups.

For experiment 1, we see that due to the simple static obstacles, the model was able to reconstruct the ground truth map state accurately. There are some errors, such as some obstacles being omitted due to occlusion from other obstacles. For experiment 2 we see similar results, but the prediction is slightly cleaner because the square robot motion path resulted in less overall occlusion of map areas. Additionally we see another omitted obstacle due to occlusion in the top middle.

For experiment 3, with the introduction of dynamic obstacles, we see a sudden drop in specificity, and the predicted map is much more blurry. Interestingly, previous work \cite{catellani2023probabilistic} modeled dynamic agent locations using probability distributions that were shaped based on limited FOV sensor input, and in our results we see a similar emergent behavior as the uncertainty introduced with obstacle motion produces a marked shift in predictions from specific location predictions to blurry probabilistic predictions. However, in our case, the probabilistic output is directly learned from the data instead of shaped using handcrafted methods. As shown in the experiment 3 prediction, the specificity of predictions correspond to the recency of sensor data, as well as the amount of occlusions. The obstacle in the top left is particularly blurry, due to its older data and larger amount of occlusion.

Similarly for experiment 4, we see blurry probabilistic predictions, however because of the robot's motion in a square shape instead of a static rotation, the sensor data for some areas of the map was much older than for others, and the recorded gradients were not as clean (e.g. scattered data in bottom). This produced a noticeable difference in the predictions for different map areas based on the recency of the sensor data and amount of occlusions from obstacle motion. We also see hallucinations in some areas.

\subsection{Ablation Study}

\begin{table*}[htb]
\centering
% \footnotesize
\caption{Prediction metrics for the ablation study where the gray-scale time decay is removed from the dynamic sensor data representation. The results for most metrics drop by 1-2\% compared to the baseline, while the average specificity drops by 9\% for exp. 3 and 12\% for exp. 4.}
\label{table:ablation}
\begin{tabular}{@{}l|llllll|llllll@{}}
\toprule
\multicolumn{1}{c|}{\textbf{DYNAMIC}} &
  \multicolumn{6}{c|}{\textbf{Experiment 3 - No Gray-Scale Time Decay}} &
  \multicolumn{6}{c}{\textbf{Experiment 4 - No Gray-Scale Time Decay}} \\ \midrule
\textbf{Model} &
  \textbf{SSIM} &
  \textbf{Acc.} &
  \textbf{Dice} &
  \textbf{Prec.} &
  \textbf{Recall} &
  \textbf{Spec.} &
  \textbf{SSIM} &
  \textbf{Acc.} &
  \textbf{Dice} &
  \textbf{Prec.} &
  \textbf{Recall} &
  \textbf{Spec.} \\ \midrule
U-Net &
  0.6493 &  0.9384 &  0.9668 &  \textbf{0.9563} &  0.9775 &  \textbf{0.5089} &  0.6493 &  \textbf{0.9340} &  0.9644 &  \textbf{0.9526} &  0.9765 &  \textbf{0.4738} \\ \midrule
FPN &
  0.6468 &  0.9392 &  0.9674 &  0.9516 &  \textbf{0.9837} &  0.4499 &  0.6396 &  0.9329 &  0.9639 &  0.9502 &  0.9779 &  0.4451 \\ \midrule
UPerNet &
  \textbf{0.6465} &  \textbf{0.9395} &  \textbf{0.9675} &  0.9540 & 0.9814 &  0.4789 &  \textbf{0.6536} &  0.9336 & \textbf{0.9644} &  0.9453 &  \textbf{0.9844} &  0.3835 \\ \midrule
Segformer &
0.6532 &  0.9373 &  0.9664 &  0.9499 &  0.9835 &  0.4293 &  0.6420 &  0.9310 &  0.9629 &  0.9481 &  0.9781 &  0.4210 \\ \midrule
\textbf{Average:} &
  \textbf{0.6490} &
  \textbf{0.9386} &
  \textbf{0.9670} &
  \textbf{0.9530} &
  \textbf{0.9815} &
  \textbf{0.4668} &
  \textbf{0.6461} &
  \textbf{0.9329} &
  \textbf{0.9639} &
  \textbf{0.9491} &
  \textbf{0.9792} &
  \textbf{0.4309} \\ \bottomrule
\end{tabular}
\end{table*}

To examine the impact of our method of cumulative dynamic sensor projection on predicted map states, we conducted an ablation study by removing the gray-scale time decay from the sensor data representations for the two dynamic experiments.

Table \ref{table:ablation} shows the results for the models run on the data without gray-scale time decay. The results for most metrics drop by 1-2\% compared to the baseline, while specificity drops by 9\% for exp. 3 and 12\% for exp. 4. This shows how the gray-scale time gradients are crucial for specific location predictions in dynamic scenarios.

Figure \ref{fig:results} shows the predicted maps for the ablation study. Compared to the baseline, we can see that the models for both experiments still capture the overall structure of the map state, but with many more errors and generally more blurry predictions. Without the gray-scale time decay there is no way to know which direction the objects have been moving in, since there is no gradient to show which sensor values are older than others. Additionally, the time since objects have last been sensed is also lost, impacting the relative confidence of different predictions, so these results match with the expected values.

%% file: 5.Conclusion.tex
\section{Conclusion}

In this work, we proposed that limited FOV sensor data can be effectively used to predict dynamic map states using existing deep learning methods. We accomplished this through a novel sensor data representation that captured sensor readings and time information in a simple single-image format, allowing for easy application of existing image-to-image prediction models. We verified the effectiveness of the proposed cumulative dynamic sensor projection approach through four different experimental setups, showing how the resulting predictions varied as obstacle and robot dynamics were changed, but the overall approach remained effective for all setups and for eight model variations. Additionally, we saw emergent predictive behavior similar to handcrafted probability distribution shaping methods from previous works, that instead occurred naturally as a result of learning from dynamic data. Through an ablation study, we verified the impact of gray-scale time decay in the cumulative dynamic sensor projection for predicting dynamic maps, and showed that the gradients produced by gray-scale time decay are crucial for predicting specific object states. These results verify that this approach can be effective in a variety of contexts, and has potential to increase the reliability of autonomous systems that utilize limited FOV sensors.

%% file: ref.bib
@INPROCEEDINGS{katyal2019uncertainty,
  author={Katyal, Kapil and Popek, Katie and Paxton, Chris and Burlina, Phil and Hager, Gregory D.},
  booktitle={2019 International Conference on Robotics and Automation (ICRA)}, 
  title={Uncertainty-Aware Occupancy Map Prediction Using Generative Networks for Robot Navigation}, 
  year={2019},
  volume={},
  number={},
  pages={5453-5459},
  doi={10.1109/ICRA.2019.8793500}}

@article{katyal2018occupancy,
  author       = {Kapil D. Katyal and
                  Katie M. Popek and
                  Chris Paxton and
                  Joseph L. Moore and
                  Kevin C. Wolfe and
                  Philippe Burlina and
                  Gregory D. Hager},
  title        = {Occupancy Map Prediction Using Generative and Fully Convolutional
                  Networks for Vehicle Navigation},
  journal      = {CoRR},
  volume       = {abs/1803.02007},
  year         = {2018},
  url          = {http://arxiv.org/abs/1803.02007},
  eprinttype    = {arXiv},
  eprint       = {1803.02007},
  bibsource    = {dblp computer science bibliography, https://dblp.org}
}

@INPROCEEDINGS{katyal2021navigation,
  author={Katyal, Kapil D. and Polevoy, Adam and Moore, Joseph and Knuth, Craig and Popek, Katie M.},
  booktitle={2021 IEEE International Conference on Robotics and Automation (ICRA)}, 
  title={High-Speed Robot Navigation using Predicted Occupancy Maps}, 
  year={2021},
  volume={},
  number={},
  pages={5476-5482},
  doi={10.1109/ICRA48506.2021.9561034}}

@article{catellani2023probabilistic,
	author={Catellani, M. and Sabattini, L.},
	journal={IEEE Robotics and Automation Letters},
	title={Distributed Control of a Limited Angular Field-of-View Multi-Robot System in Communication-Denied Scenarios: A Probabilistic Approach},
	year={2024},
	volume={9},
	number={1},
	pages={739-746},
	doi={10.1109/LRA.2023.3337694},
	ISSN={},
	month={January}
}

@ARTICLE{nguyen2021distributed,
  author={Nguyen, Hoa Van and Rezatofighi, Hamid and Vo, Ba-Ngu and Ranasinghe, Damith C.},
  journal={IEEE Transactions on Signal Processing}, 
  title={Distributed Multi-Object Tracking Under Limited Field of View Sensors}, 
  year={2021},
  volume={69},
  number={},
  pages={5329-5344},
  doi={10.1109/TSP.2021.3103125}}

@ARTICLE{mukherjee2022distributed,
  author={Mukherjee, Pratik and Santilli, Matteo and Gasparri, Andrea and Williams, Ryan K.},
  journal={IEEE Robotics and Automation Letters}, 
  title={Distributed Adaptive and Resilient Control of Multi-Robot Systems With Limited Field of View Interactions}, 
  year={2022},
  volume={7},
  number={2},
  pages={5318-5325},
  doi={10.1109/LRA.2022.3155822}}

@INPROCEEDINGS{sung2017searching,
  author={Sung, Yoonchang and Tokekar, Pratap},
  booktitle={2017 IEEE International Conference on Robotics and Automation}, 
  title={Algorithm for searching and tracking an unknown and varying number of mobile targets using a limited FoV sensor}, 
  year={2017},
  volume={},
  number={},
  pages={6246-6252},
  doi={10.1109/ICRA.2017.7989740}}

@inproceedings{xie2023sogmp,
  doi = {10.48550/ARXIV.2210.08577},
  title={Stochastic Occupancy Grid Map Prediction in Dynamic Scenes},
  author={Zhanteng Xie and Philip Dames},
  booktitle={Proceedings of The 7th Conference on Robot Learning},
  pages={1686--1705},
  year={2023},
  volume={229},
  series={Proceedings of Machine Learning Research},
  month={06--09 Nov},
  publisher={PMLR},
  url={https://proceedings.mlr.press/v229/xie23a.html}
}

@misc{segmodels,
  Author = {Pavel Iakubovskii},
  Title = {Segmentation Models Pytorch},
  Year = {2019},
  Publisher = {GitHub},
  Journal = {GitHub repository},
  Howpublished = {\url{https://github.com/qubvel/segmentation\_models.pytorch}}
}

@misc{piqa,
  Author = {François Rozet},
  Title = {PyTorch Image Quality Assessment},
  Year = {2020},
  Publisher = {GitHub},
  Journal = {GitHub repository},
  Howpublished = {\url{https://github.com/francois-rozet/piqa/tree/master}}
}

@misc{segmetrics,
  Author = {Hsiangyu Zhao},
  Title = {Segmentation Metrics Pytorch},
  Year = {2020},
  Publisher = {GitHub},
  Journal = {GitHub repository},
  Howpublished = {\url{https://github.com/hsiangyuzhao/Segmentation-Metrics-PyTorch}}
}

@misc{irsim,
  Author = {Han Ruihua},
  Title = {Intelligent Robot Simulator},
  Year = {2022},
  Publisher = {GitHub},
  Journal = {GitHub repository},
  Howpublished = {\url{https://github.com/hanruihua/ir-sim}}
}

@INPROCEEDINGS{sharma2023,
  author={Sharma, Vishnu D. and Chen, Jingxi and Tokekar, Pratap},
  booktitle={2023 IEEE/RSJ International Conference on Intelligent Robots and Systems (IROS)}, 
  title={ProxMaP: Proximal Occupancy Map Prediction for Efficient Indoor Robot Navigation}, 
  year={2023},
  volume={},
  number={},
  pages={7135-7140},
  doi={10.1109/IROS55552.2023.10341435}
}

@INPROCEEDINGS{shrestha2019,
  author={Shrestha, Rakesh and Tian, Fei-Peng and Feng, Wei and Tan, Ping and Vaughan, Richard},
  booktitle={2019 International Conference on Robotics and Automation (ICRA)}, 
  title={Learned Map Prediction for Enhanced Mobile Robot Exploration}, 
  year={2019},
  volume={},
  number={},
  pages={1197-1204},
  doi={10.1109/ICRA.2019.8793769}
}

@INPROCEEDINGS{xia2024,
  author={Xia, Xingyu and Zhu, Hai and Zhu, Xiaozhou and Yao, Wen},
  booktitle={2024 IEEE International Conference on Unmanned Systems (ICUS)}, 
  title={Learning Predicted Occupancy Map for Risk-Aware MAV Motion Planning in Dynamic Environments}, 
  year={2024},
  volume={},
  number={},
  pages={1654-1659},
  doi={10.1109/ICUS61736.2024.10840031}
}

@INPROCEEDINGS{doherty2017,
  author={Doherty, Kevin and Wang, Jinkun and Englot, Brendan},
  booktitle={2017 IEEE International Conference on Robotics and Automation}, 
  title={Bayesian generalized kernel inference for occupancy map prediction}, 
  year={2017},
  volume={},
  number={},
  pages={3118-3124},
  doi={10.1109/ICRA.2017.7989356}
}

@inproceedings{unet,
author = {Ronneberger, Olaf and Fischer, Philipp and Brox, Thomas},
year = {2015},
month = {10},
pages = {234-241},
title = {U-Net: Convolutional Networks for Biomedical Image Segmentation},
volume = {9351},
isbn = {978-3-319-24573-7},
journal = {LNCS},
doi = {10.1007/978-3-319-24574-4_28}
}

@INPROCEEDINGS{fpn,
  author={Lin, Tsung-Yi and Dollár, Piotr and Girshick, Ross and He, Kaiming and Hariharan, Bharath and Belongie, Serge},
  booktitle={2017 IEEE Conference on Computer Vision and Pattern Recognition (CVPR)}, 
  title={Feature Pyramid Networks for Object Detection}, 
  year={2017},
  volume={},
  number={},
  pages={936-944},
  doi={10.1109/CVPR.2017.106}
}

@inproceedings{upernet,
  title={Unified Perceptual Parsing for Scene Understanding},
  author={Xiao, Tete and Liu, Yingcheng and Zhou, Bolei and Jiang, Yuning and Sun, Jian},
  booktitle={European Conference on Computer Vision},
  year={2018},
  organization={Springer}
}

@inproceedings{xie2021segformer,
  title={SegFormer: Simple and Efficient Design for Semantic Segmentation with Transformers},
  author={Xie, Enze and Wang, Wenhai and Yu, Zhiding and Anandkumar, Anima and Alvarez, Jose M and Luo, Ping},
  booktitle={Neural Information Processing Systems (NeurIPS)},
  year={2021}
}

@INPROCEEDINGS{hoermann2018dynamic,
  author={Hoermann, Stefan and Bach, Martin and Dietmayer, Klaus},
  booktitle={2018 IEEE International Conference on Robotics and Automation (ICRA)}, 
  title={Dynamic Occupancy Grid Prediction for Urban Autonomous Driving: A Deep Learning Approach with Fully Automatic Labeling}, 
  year={2018},
  volume={},
  number={},
  pages={2056-2063},
  doi={10.1109/ICRA.2018.8460874}
}

@INPROCEEDINGS{guizilini2019hilbert,
  author={Guizilini, Vitor and Senanayake, Ransalu and Ramos, Fabio},
  booktitle={2019 International Conference on Robotics and Automation (ICRA)}, 
  title={Dynamic Hilbert Maps: Real-Time Occupancy Predictions in Changing Environments}, 
  year={2019},
  volume={},
  number={},
  pages={4091-4097},
  doi={10.1109/ICRA.2019.8793914}
}
